\documentclass{article}

\usepackage[square,sort,comma,numbers]{natbib}

\usepackage[preprint]{neurips_2023}

\usepackage[utf8]{inputenc} 
\usepackage[T1]{fontenc}    
\usepackage{hyperref}       
\usepackage{url}            
\usepackage{booktabs}       
\usepackage{amsfonts}       
\usepackage{nicefrac}       
\usepackage{microtype}      
\usepackage{xcolor}         
\usepackage{graphicx}
\usepackage{subfig}
\usepackage{multirow}
\usepackage{wrapfig}
\usepackage{amsmath}
\usepackage{physics}
\usepackage{adjustbox}
\usepackage[]{hyperref} 
\hypersetup{
colorlinks=true,
}

\title{Long Short-term Memory with \\ Two-Compartment Spiking Neuron}

\author{%
  Shimin Zhang$^1$\thanks{Equal contribution}, Qu Yang$^2$\footnotemark[1], Chenxiang Ma$^1$, Jibin Wu$^1$\thanks{Corresponding Author: jibin.wu@polyu.edu.hk}, Haizhou Li$^{2,3}$, Kay Chen Tan$^1$\\
  $^1$The Hong Kong Polytechnic University\\
  $^2$National University of Singapore\\
  $^3$The Chinese University of Hong Kong, Shenzhen, China\\
}

\begin{document}

\maketitle

\begin{abstract}
The identification of sensory cues associated with potential opportunities and dangers is frequently complicated by unrelated events that separate useful cues by long delays. As a result, it remains a challenging task for state-of-the-art spiking neural networks (SNNs) to identify long-term temporal dependencies since bridging the temporal gap necessitates an extended memory capacity. To address this challenge, we propose a novel biologically inspired \textbf{L}ong \textbf{S}hort-\textbf{T}erm \textbf{M}emory \textbf{L}eaky \textbf{I}ntegrate-and-\textbf{F}ire spiking neuron model, dubbed LSTM-LIF. Our model incorporates carefully designed somatic and dendritic compartments that are tailored to retain short- and long-term memories. The theoretical analysis further confirms its effectiveness in addressing the notorious vanishing gradient problem. Our experimental results, on a diverse range of temporal classification tasks, demonstrate superior temporal classification capability, rapid training convergence, strong network generalizability, and high energy efficiency of the proposed LSTM-LIF model. This work, therefore, opens up a myriad of opportunities for resolving challenging temporal processing tasks on emerging neuromorphic computing machines.

\end{abstract}

\section{Introduction}
\label{sec: intro}
Deep learning has revolutionized many research fields by empowering machines to learn complex patterns from vast amounts of data, instances include computer vision \cite{krizhevsky2017imagenet, tan2020efficientdet, he2022masked}, natural language processing \cite{devlin2018bert, radford2019language, brown2020language}, and speech recognition \cite{graves2013speech, gulati2020conformer, he2019streaming}. A key component of deep learning is the artificial neural network, which is inspired by the structure and function of biological neural networks \cite{lecun2015deep}. Among the different types of artificial neural networks, spiking neural networks (SNNs) have attracted significant attention recently owing to their biological plausibility and potential for facilitating energy-efficient computation \cite{maass1997networks, furber2014spinnaker, pfeiffer2018deep}.

Spiking neurons emulate the rich neuronal dynamics of biological neurons, which facilitate the encoding and memorizing of spatio-temporal sensory cues. Furthermore, spiking neurons communicate with each other via discrete spikes, such event-driven operation leads to ultra-low-power neural computation \cite{gerstner2014neuronal, pfeiffer2018deep}. 
In practice, single-compartment spiking neurons models are widely adopted due to their mathematical tractability and computational efficiency, instances include Leaky Integrate-and-Fire (LIF) Model \cite{abbott2005model}, Izhikevich Model \cite{izhikevich2003simple}, and Adaptive Exponential Integrate-and-Fire (AdEx) Model \cite{brette2005adaptive}. These single-compartment models abstract the biological neuron as a single electrical circuit, preserving the essential neuronal dynamics of biological neurons while ignoring the complex geometrical structure of dendrites and somas. This degree of abstraction significantly reduces the modeling effort, making them more feasible to study the behavior of large-scale biological neural networks and perform complex pattern recognition tasks on neuromorphic machines.

While single-compartment spiking neuron models have demonstrated promising results in various pattern recognition tasks \cite{tavanaei2019deep, zhang2021rectified, wu2021progressive, bu2023optimal, yang2022training, yao2022glif}, their ability to solve tasks that require long-term temporal dependencies remains constrained. This is primarily attributed to their limited memory capacity. Specifically, the intrinsic leakage of neuronal states (e.g., membrane potential) coupled with the reset mechanism lead to rapid forget or loss of past inputs  \cite{gerstner2014neuronal}. The loss of information about past inputs makes it challenging for neurons to learn long-term dependencies, especially when the temporal gap between sensory cues is significantly larger than the decaying time constant of the neuronal state variables \cite{maass2002model}. This problem has motivated the recent proposals of adopting dynamic firing threshold \cite{bellec2018long,yin2020effective, yin2021accurate} and adaptive time constant \cite{yin2020effective, yin2021accurate,fang2021incorporating} to improve the memory capacity of single-compartment spiking neurons.

For biological neurons, the separation of dendrites and soma as well as the 
the complex geometrical structure of dendrites facilitates interactions between different neuronal compartments, resulting in memory traces of input signals at different timescales \cite{stuart2015dendritic}. This multi-compartment structure enhances neuronal dynamics and allows historical or contextual information to be maintained over an extended time period. It, therefore, lays the foundation for learning long-term dependencies between sensory cues \cite{london2005dendritic}. While incorporating more compartments offers additional benefits of expanded memory capacity, the increased model complexity and computational cost may hinder their practical use, especially for complex pattern recognition tasks with large-scale SNNs. 

In this paper, we derive a generalized two-compartment neuron model as depicted in Figure \ref{fig: inter_ops}(a). This neuron model provides an ideal reflection of the minimal geometry of the well-known Prinsky-Rinzel (P-R) pyramidal neuron while preserving the essential features of more complicated multi-compartment models \cite{lin2017dynamical}. Furthermore, we proposed a memory-augmented variant, which we referred to as \textbf{L}ong \textbf{S}hort-\textbf{T}erm \textbf{M}emory \textbf{L}eaky \textbf{I}ntegrate-and-\textbf{F}ire (LSTM-LIF) model. LSTM-LIF separates soma and dendrites into two compartments, which are tailored to store short-term and long-term memories, respectively. This unique design significantly boosts the memory capacity of traditional single-compartment neuron models, thereby enabling the effective processing of multi-scale temporal information.

The main contributions of our work are summarized as follows:
\begin{itemize}
\item 
We propose a biologically inspired two-compartment spiking neuron model, dubbed LSTM-LIF, which tailors its somatic and dendritic compartments to store short- and long-term memories, respectively.
\item We conduct a theoretical analysis to shed light on the effectiveness of our proposed LSTM-LIF model in resolving the vanishing gradient problem during BPTT training.
\item Our experimental results, on a broad range of temporal classification tasks, demonstrate the superior performance of the proposed model, including exceptional classification capability, rapid training convergence, greater network generalizability, and high energy efficiency. 
\end{itemize}

\section{Related works}
\label{sec: related_works}

\subsection{Memory-Enhanced Single-Compartment Spiking Neuron Models}
\label{sec: enhanced_single}
Due to the inherent limitations of LIF neurons in performing long-term temporal credit assignments, the development of memory-enhanced single-compartment spiking neuron models has become a research focus in recent years. Notable efforts include the Long Short-Term Memory Spiking Neural Network (LSNN) proposed by Bellec et al. \cite{bellec2018long}. LSNN introduces an adaptive firing threshold mechanism to LIF neurons, which serves as a long-term memory of past inputs. Yin et al. \cite{yin2020effective, yin2021accurate} further propose to apply learnable time constants for the adaptive firing threshold, such that multi-scale temporal information can be retained \cite{yin2020effective, yin2021accurate}. 
Along the same direction, the Parametric LIF (PLIF) model \cite{fang2021incorporating} introduces learnable membrane time constants that allow LIF neurons to retain multi-scale temporal information in their membrane variables. More recently, the gated LIF (GLIF) \cite{yao2022glif} model incorporates learnable gates to selectively integrate the essential neuronal dynamics, including synaptic integration, membrane leakage, and reset. These enhancements enrich the neuronal dynamics and, therefore, improve the representation power and adaptivity of LIF neurons. 
Nevertheless, these single-compartment spiking neuron models are still 
facing problems with limited memory capacity and struggle to perform long-term temporal credit assignments. This motivates us to explore more complex multi-compartment models in this work to further enhance the memory capacity of spiking neurons.

\subsection{Multi-Compartment Spiking Neuron Model}
\label{sec: multi}
Multi-compartment spiking neuron models have been extensively studied in literatures. By faithfully modeling the geometrical structure of biological neurons as well as the interactions among their different compartments, multi-compartment models can represent the rich neuronal dynamics of biological neurons. One of the earliest multi-compartment models is the Rall model \cite{rall1964theoretical}, which is designed based on the cable theory of passive dendrites.
While the Rall model is primarily focused on passive dendritic properties, other models have been extended to incorporate active conductances, such as voltage-gated ion channels, that play a crucial role in shaping the dendritic voltage responses. For instance, the Pinsky-Rinzel model \cite{pinsky1994intrinsic} is a two-compartment model that simulates the interaction between somatic and dendritic compartments, capturing the essential properties of CA3 pyramidal neurons in the hippocampus. 
More recently, researchers have proposed multi-compartment models with varying levels of complexity to better understand the role of dendrites in neural computation. 
Multi-compartment spiking neuron models have proven valuable in understanding the complex dynamics of biological neurons, as well as in enabling more accurate brain simulations. However, the trade-off between model complexity and computational efficiency remains a key challenge for practical applications. In this work, we aim to design a biologically plausible two-compartment neuron model that can achieve a good balance between these two aspects, while maintaining the superior temporal processing capability of biological neurons.

\section{Methodology}
\label{sec: methodology}

In this section, we first introduce the dynamics of a typical single-compartment neuron model (i.e., LIF) and elaborate on its inherent deficiencies in retaining long-term memories as well as learning long-term dependencies. Then, we will present a generalized two-compartment spiking neuron model inspired by the well-known Prinsky-Rinzel pyramidal neurons \cite{pinsky1994intrinsic}. Based on this, we further develop a memory-augmented two-compartment spiking neuron model, namely LSTM-LIF. 
Our proposed LSTM-LIF model is theoretically well-grounded that can facilitate learning long-term dependencies.

\subsection{Inherent Limitations of LIF Neuron for Long-term Temporal Credit Assignment}
\label{sec: SNN}

In general, spiking neurons integrate synaptic inputs triggered by incoming spikes. Once the accumulated membrane potential surpasses the firing threshold, an output spike will be generated and transmitted to subsequent neurons. The  LIF neuron is the most ubiquitous and effective single-compartment spiking neuron model, which has been widely used for large-scale brain simulation and neuromorphic computing. The neuronal dynamics of a LIF neuron can be described by the following discrete-time formulations:

\begin{equation}
\label{LIF_dynamics_whole}
\mathcal{U}[t]=\beta\hspace{0.5mm}\mathcal{U}[t-1] - \mathcal{V}_{th}\mathcal{S}[t-1] + \mathcal{I}[t] 
\end{equation}

\begin{equation}
\label{Inj_I}
\mathcal{I}[t]=\sum_{i}{\omega_{i}\mathcal{S}_{i}[t-1]}+b 
\end{equation}

\begin{equation}
\label{Surrogate_spike_cal}
\mathcal{S}[t]=\Theta(\mathcal{U}[t]-\mathcal{V}_{th})
\end{equation}

\begin{figure}[htb]
\centering
\subfloat[]{
\begin{adjustbox}{raise=0.5cm}
    \includegraphics[width=45mm]{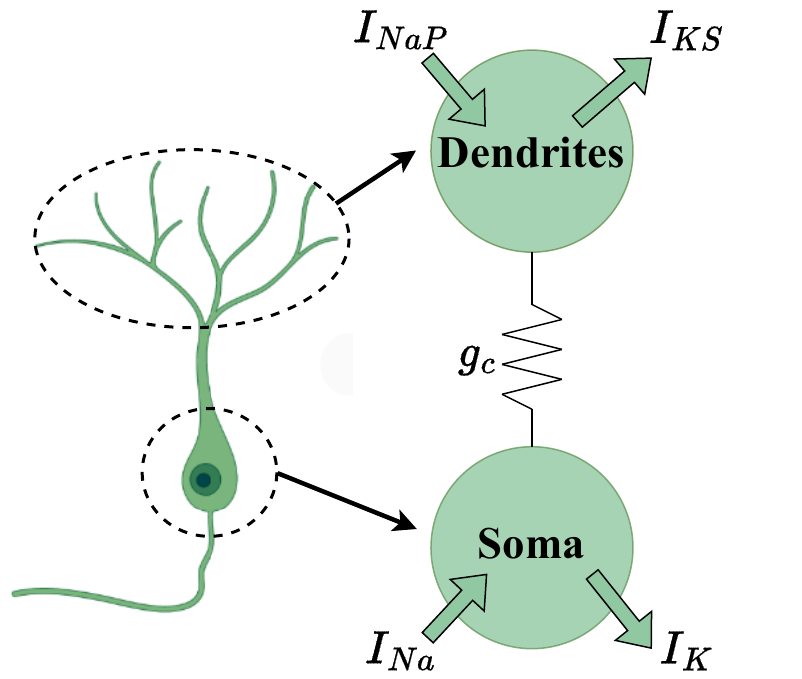}
\end{adjustbox}
} \quad
\subfloat[]{\includegraphics[width=40mm]{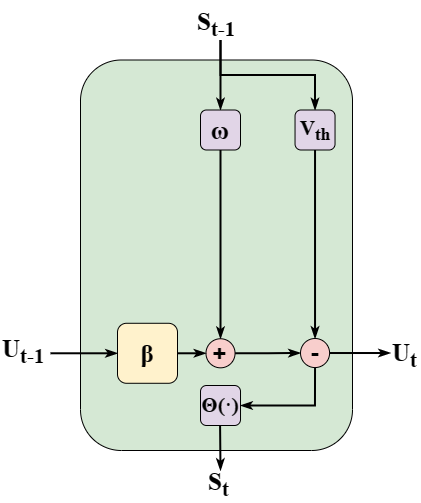}} \quad \quad 
\subfloat[]{\includegraphics[width=40mm]{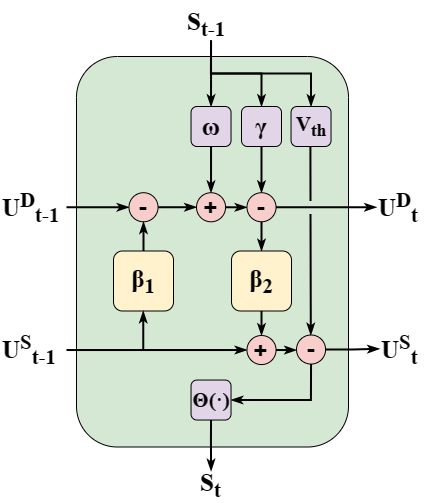}} 
\caption{Illustration of (b) the structure of a two-compartment Prinsky-Rinzel pyramidal neuron, and the internal operations of (b) LIF model as well as the proposed (c) LSTM-LIF model.}
\label{fig: inter_ops}
\end{figure}

where $\mathcal{U}[t]$ and $\mathcal{I}[t]$ represent the membrane potential and the input current of a neuron at time $t$, respectively. The term $\beta \equiv exp(-dt/\tau_{m})$ is the membrane decaying coefficient that ranged from (0, 1), in which $\tau_m$ is the membrane time constant and $dt$ is the simulation time step. $\omega_{i}$ denotes the synaptic weight that connects input neuron $i$, and $b$ represents the bias term. An output spike will be generated once the membrane potential $\mathcal{U}[t]$ crosses the neuronal firing threshold $\mathcal{V}_{th}$ as per Eq. (\ref{Surrogate_spike_cal}).

The vanishing gradient problem remains a critical obstacle that hampers the learning of long-term dependencies by stateful neural networks, such as vanilla RNNs and SNNs. 
To further elaborate on this problem in SNN, we consider the network training with the following objective function:

\begin{equation}
\label{SNN_obj}
\mathcal{L}(\hat{\mathcal{S}}, \mathcal{S})=\frac{1}{N}\sum_{n=1}^{N}{\mathcal{L}}(\hat{\mathcal{S}}_{n}, \mathcal{S}_{n})
\end{equation}

where $N$ is the number of training samples, ${\mathcal{L}}$ is the loss function, ${\mathcal{S}}_{n}$ is the network output, and $\hat{\mathcal{S}}_{n}$ is the training target.
This objective function can be optimized with the canonical backpropagation through time (BPTT) algorithm. In particular, the gradient of the synaptic weight $\omega$ can be calculated as follows:

\begin{equation}
\label{SNN_gradient}
\frac{\partial\mathcal{L}}{\partial\omega}=\sum_{t=1}^{T}\frac{\partial\mathcal{L}}{\partial\mathcal{S}[T]}\frac{\partial\mathcal{S}[T]}{\partial\mathcal{U}[T]}\frac{\partial\mathcal{U}[T]}{\partial\mathcal{U}[t]}\frac{\partial\mathcal{U}[t]}{\partial\omega}
\end{equation}

By substituting Eq. (\ref{LIF_dynamics_whole}) into the above equation to compute $\partial\mathcal{U}{[T]}/\partial\mathcal{U}{[t]}$, it is obvious that the influence of time step $t$ on its subsequent time step $T$ diminish when $T$ increases. This is because the membrane potential decay causes an exponential decay of early information. This problem becomes exacerbated when $t$ is considerably smaller than $T$, leading to the vanishing gradient problem.
Consequently, single-compartment neuron models, epitomized by the LIF model, are struggled to retain long-term memory. Therefore, their ability to learn long-term dependencies are limited. This motivates us to develop two-compartment neuron models that can expand memory capacity and facilitate learning long-term dependencies.

\subsection{Generalized Two-Compartment Spiking Neuron Model}
\label{sec: PR_model}
The Prinsky-Rinzel (P-R) pyramidal neurons are located in the CA3 region of the hippocampus, which plays an important role in memory storage and retrieval of animals. Researchers have simplified this neuron model as a two-compartment model that can simulate the interaction between somatic and dendritic compartments, as depicted in Figure \ref{fig: inter_ops}(a). Drawing upon the structure of the P-R model, we develop a generalized two-compartment spiking neuron model that defined as the following. The detailed derivations of this general formulation are provided in Supplementary Material Section \ref{app: method}. 

\begin{equation}
\label{PR_dendrite_general}
\mathcal{U}^{D}[t]=\alpha_1\hspace{0.5mm}\mathcal{U}^{D}[t-1] + \beta_1\hspace{0.5mm}\mathcal{U}^{S}[t-1] + \mathcal{I}[t]
\end{equation}

\begin{equation}
\label{PR_soma_general}
\mathcal{U}^{S}[t]=\alpha_2\hspace{0.5mm}\mathcal{U}^{S}[t-1] + \beta_2\hspace{0.5mm}\mathcal{U}^{D}[t] - \mathcal{V}_{th}\mathcal{S}[t-1]
\end{equation}

\begin{equation}
\label{PR_spike_cal}
\mathcal{S}[t]=\Theta(\mathcal{U}^{S}[t]-\mathcal{V}_{th})
\end{equation}

where $\hspace{0.5mm}\mathcal{U}^{D}$ and $\hspace{0.5mm}\mathcal{U}^{S}$ represents the membrane potentials of the dendritic and the somatic compartments, respectively. 
$\alpha_1$ and $\alpha_2$ are respective membrane potential decaying coefficients for these two compartments. 
Notably, the membrane potentials of these two compartments are not updated independently. Rather, they are coupled with each other through the second term in Eqs. (\ref{PR_dendrite_general}) and (\ref{PR_soma_general}), in which the coupling effects are controlled by the coefficients $\beta_1$ and $\beta_2$.
The interplay between these two compartments enhances the neuronal dynamics and, if properly designed, can resolve the vanishing gradient problem.

\subsection{Long Short-Term Memory Leaky Integrate-and-Fire (LSTM-LIF) Model}
\label{sec: TCSN}

Based on the generalized two-compartment spiking neuron model introduced earlier, we propose an LSTM-LIF model that equips with enhanced memory capacity as well as the ability to learn long-term dependencies. In comparison to the generalized two-compartment neuron model, we drop the membrane decaying factors $\alpha_1$ and $\alpha_1$ from both compartments. This modification aims to circumvent the rapid decay of memory that could cause unintended information loss. Moreover, to circumvent excess firing caused by persistent input accumulation, we design $\beta_1$ and $\beta_2$ to take opposite signs. The dynamics of the proposed LSTM-LIF model are defined by the following equations: 
\begin{equation}
\label{TCME_dendrite}
\mathcal{U}^{D}[t]=\mathcal{U}^{D}[t-1] + \beta_1\hspace{0.5mm}\mathcal{U}^{S}[t-1] + \mathcal{I}[t] - \gamma\mathcal{S}[t-1]
\end{equation}

\begin{equation}
\label{TCME_soma}
\mathcal{U}^{S}[t]=\mathcal{U}^{S}[t-1] + \beta_2\hspace{0.5mm}\mathcal{U}^{D}[t] - \mathcal{V}_{th}\mathcal{S}[t-1] 
\end{equation}

\begin{equation}
\label{TCME_spike_cal}
\mathcal{S}[t]=\Theta(\mathcal{U}^{S}[t]-\mathcal{V}_{th})
\end{equation}

According to the above formulations, $\hspace{0.5mm}\mathcal{U}^{S}$ is responsible for retaining short-term memory about decaying dendritic inputs, which will be reset after neuron firing. Notably, the output spikes are generated from the somatic compartment in a context-aware manner that contributed by $\mathcal{U}^{D}$. In contrast, $\hspace{0.5mm}\mathcal{U}^{D}$ serves as a long-term memory that retains the past inputs. It is worth noting that despite negative feedback from the somatic compartment, the memory traces of $\mathcal{I}[t]$ will not be corrupted.  
The coefficients $\beta_1 \equiv -\sigma(c_1)$ and $\beta_2 \equiv \sigma(c_2)$ determine the efficacy of information communication between two compartments. Here, the sigmoid function $\sigma(\cdot)$ is utilized to ensure two coefficients are within the range of (-1, 0) and (0, 1), and the parameters $c_1$ and $c_2$ can be automatically adjusted during the training process. The effect of this design choice will be analyzed in details in Section \ref{sec: beta_study}. The membrane potentials of both compartments are reset after the firing of the soma. The reset of the dendritic compartment is triggered by the backpropagating spike that governed by a scaling factor $\gamma$. The internal operations of the LSTM-LIF model are depicted in Figure \ref{fig: inter_ops}(c), which exhibits richer internal dynamics in comparison to the LIF model that is shown in Figure \ref{fig: inter_ops}(b).

To further demonstrate the superiority of the proposed  model in learning long-term dependencies, we provide a mathematical proof to show why the LSTM-LIF model can greatly alleviate the vanishing gradient problem. 
As discussed in Section \ref{sec: SNN}, the primary cause of the gradient vanishing problem is attributed to the recursive computation of $\partial\mathcal{U}_{T}/\partial\mathcal{U}_{t}$.
This problem can, however, be effectively alleviated in the proposed LSTM-LIF model, wherein the partial derivative $\partial\mathcal{U}_{T}/\partial\mathcal{U}_{t}$ can be calculated as follows:
\begin{equation}
\label{TCME_adj_gradient_aggregation}
\frac{\partial\hspace{0.5mm}\mathcal{U}[T]}{\partial\hspace{0.5mm}\mathcal{U}[t]}=\prod_{j=t+1}^{T}\frac{\partial\hspace{0.5mm}\mathcal{U}[j]}{\partial\hspace{0.5mm}\mathcal{U}[j-1]}, \hspace{1.0mm}\mathcal{U}[j]=\left[\mathcal{U}^{D}[j], \mathcal{U}^{S}[j]\right]^T
\end{equation}

where
\begin{equation}
\label{TCME_adj_gradient_matrice}
\frac{\partial\hspace{0.5mm}\mathcal{U}[j]}{\partial\hspace{0.5mm}\mathcal{U}[j-1]}=
\begin{bmatrix}
    \frac{\partial\hspace{0.5mm}\mathcal{U}^{D}[j]}{\partial\hspace{0.5mm}\mathcal{U}^{D}[j-1]} & \frac{\partial\hspace{0.5mm}\mathcal{U}^{D}[j]}{\partial\hspace{0.5mm}\mathcal{U}^{S}[j-1]} \\
    \\[-1ex]
    \frac{\partial\hspace{0.5mm}\mathcal{U}^{S}[j]}{\partial\hspace{0.5mm}\mathcal{U}^{D}[j-1]} & \frac{\partial\hspace{0.5mm}\mathcal{U}^{S}[j]}{\partial\hspace{0.5mm}\mathcal{U}^{S}[j-1]}
\end{bmatrix}=
\begin{bmatrix}
    \beta_1\beta_2+1 & \beta_1 \\
    \\[-1ex]
    \beta_1\beta_2^2+2\beta_2 & \beta_1\beta_2+1
\end{bmatrix}
\end{equation}

In order to quantify the severity of the vanishing gradient problem in LSTM-LIF, we further calculate the 
column infinite norm as provided in Eq. (\ref{TCME_adj_gradient_inf_norm}). The infinite norm signifies the maximum changing rate of membrane potentials over a prolonged time period.

\begin{equation}
\label{TCME_adj_gradient_inf_norm}
\norm{\frac{\partial\hspace{0.5mm}\mathcal{U}[j]}{\partial\hspace{0.5mm}\mathcal{U}[j-1]}}_{\infty}=\max(\beta_1\beta_2^2+\beta_1\beta_2+2\beta_2+1, \beta_1\beta_2+\beta_1+1)
\end{equation}

By employing the constrained optimization method to solve the lower bound of Eq. (\ref{TCME_adj_gradient_inf_norm}), it can be found that  $\norm{\partial\hspace{0.5mm}\mathcal{U}[j]/\partial\hspace{0.5mm}\mathcal{U}[j-1]}_{\infty}>1$. This suggests LSTM-LIF model can effectively prevent the unexpected occurrence of exponentially decaying gradients. 

It is worth noting that the LSTM-LIF model can be reformulated into a single-compartment form:

\begin{equation}
\label{TCME_single_form}
\mathcal{U}^{S}[t]=(1+\beta_1\hspace{0.5mm}\beta_2)\mathcal{U}^{S}[t-1] + \beta_2\hspace{0.5mm}\mathcal{U}^{D}[t-1] + \beta_2\hspace{0.5mm}\mathcal{I}[t] - (\beta_2\gamma+\mathcal{V}_{th})\mathcal{S}[t-1] 
\end{equation}

In essence, the above formulation mirrors a LIF neuron that characterized by a decaying input. 
Although memory decaying problem remains inextricable for the LSTM-LIF model, the presence of $\mathcal{U}^D$ can effectively compensate for the memory loss and address the vanishing gradient problem.
 
\section{Experiments}
\label{sec: exp}
In this section, we first validate the effectiveness of our proposed LSTM-LIF model in learning long-term dependencies. Then, we evaluate the LSTM-LIF model on various temporal classification benchmarks, including sequential MNIST (S-MNIST) \cite{le2015simple}, permuted sequential MNIST (PS-MNIST) \cite{le2015simple}, Google Speech Commands (GSC) \cite{warden2018speech}, Spiking Heidelberg Digits (SHD) \cite{cramer2020heidelberg}, and Spiking Google Speech Commands (SSC) \cite{cramer2020heidelberg}. 
Finally, we conduct a comprehensive study to demonstrate the advantages of the LSTM-LIF model in terms of rapid training convergence, strong network generalization, and high energy efficiency. To facilitate comparison with state-of-the-art (SOTA) single-compartment neuron models, we construct our network architectures with a comparable amount of parameters. More details about our experimental setups are provided in Supplementary Materials Section \ref{app: exp_setting}. 

\subsection{Exploring Parameter Space for Generalized Two-Compartment Neurons}
\label{sec: beta_study}

\begin{wrapfigure}{r}{0.5\textwidth}
\centering
\subfloat{\includegraphics[width=68mm]{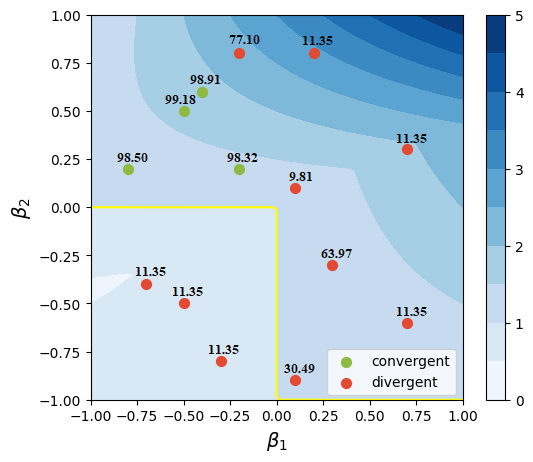}}
\caption{Model accuracies with respect to different initial beta values.}
\label{fig: beta_study}
\end{wrapfigure}

In Section \ref{sec: PR_model}, we put forward a generalized two-compartment model whose neuronal dynamics are determined by $[\alpha_1, \alpha_2, \beta_1, \beta_2]$. To circumvent the rapid decay of memory, we set both $\alpha_1$ and $\alpha_2$ to one. It is worth noting that the selection of $\beta_1$ and $\beta_2$ will, however, significantly affect the training convergence of a two-compartment neuron model. To shed light on the effectiveness of the proposed parameter setting for the LSTM-LIF model, we initialize $\beta_1$ and $\beta_2$ across four different quadrants and evaluate their performance on the S-MNIST dataset.

In Figure \ref{fig: beta_study}, the contour map illustrates the partial derivative of the membrane potential at adjacent time steps for a generalized two-compartment neuron model.  
Different color shades on this map indicate different values of the partial derivative.
The yellow line demarcates the region where the partial derivative equals one. The scattered points on this plot represent different neuron models, each initialized with a different set of values for $\beta_1$ and $\beta_2$.

The result reveals that when initializing $\beta$ in the first and the third quadrants leads to apparent exploding and vanishing gradient problems respectively, resulting in the models to diverge. While initializing $\beta$ values within the upper-right corner of the second quadrant leads to the issue of exploding gradients, a wide range of $\beta$ values within the lower-left corner of the second quadrant can support effective training. Therefore, we select initialization values from this region for our LSTM-LIF model and we use it consistently for the rest of our experiments. Although initializing $\beta$ in the fourth quadrant can avoid the issue of vanishing and exploding gradients, it results in negative inputs (see Eq. (\ref{TCME_soma})) to the somatic compartment that will lead to poor temporal classification results. 

\subsection{Superior Performance for Temporal Classification Tasks}
\label{sec: comp_sota}
Table \ref{tab: comp_sota} presents the results of the proposed LSTM-LIF model on five selected datasets, along with other existing works. Overall, given the same amount of parameters, the LSTM-LIF model consistently outperforms SOTA single-compartment neurons across all datasets.

\begin{table*}[t]
  \caption{Comparison of model performance on S-MNIST, PS-MNIST, GSC, SHD, and SSC datasets.}
  \label{tab: comp_sota}
  \centering
  \resizebox{0.9\textwidth}{!}{%
  \begin{tabular}{cllcc}
    \toprule
    \textbf{Datasets}   &\textbf{Method}    &\textbf{Network}   &\textbf{Parameters (K)}    &\textbf{Accuracy (\%)} \\
    \midrule
    \multirow{12}{*}{\rotatebox{90}{S-MNIST}}      & GLIF* \cite{yao2022glif}     & feedforward       & 47.1/87.5  & 94.80/95.27 \\               & PLIF* \cite{fang2021incorporating}    & feedforward     & 44.8/85.1  & 83.71/87.92 \\
                  & LIF*      & feedforward         & 44.8/85.1         &62.42/72.06 \\
                  & \textbf{LSTM-LIF (ours)} &  \textbf{feedforward} & \textbf{44.8/85.1}   & \textbf{96.46/97.35} \\
    \cmidrule(r){2-5}
                  & LSTM (non-spiking) \cite{arjovsky2016unitary}& recurrent         & 66.5/ -        & 98.20/ - \\
                  & SRNN+ReLU (non-spiking) \cite{yin2020effective}         & recurrent         & 129.6/ -         & 98.99/ - \\
                  & LSNN \cite{bellec2018long}              & recurrent         & 68.2/ -        & 93.70/ - \\
                  & GLIF* \cite{yao2022glif}            & recurrent       & 114.6/157.5     & 95.63/96.64 \\
                  & SRNN+ALIF \cite{yin2020effective,yin2021accurate} & recurrent         & 129.6/156.3         & 97.82/98.70 \\
                  & PLIF* \cite{fang2021incorporating}    & recurrent     & 112.2/155.1  & 90.93/91.79 \\
                  & LIF*                            & recurrent         & 112.2/155.1         & 74.91/89.28 \\
                  & \textbf{LSTM-LIF (ours)} &  \textbf{recurrent}  &  \textbf{63.6/155.1}  &  \textbf{98.79/99.18} \\
    \bottomrule
    \multirow{8}{*}{\rotatebox{90}{PS-MNIST}}   
                  & LIF*      & feedforward         & 44.8/85.1         & 11.30/10.00 \\
                  & \textbf{LSTM-LIF (ours)} &  \textbf{feedforward} & \textbf{44.8/85.1}   & \textbf{80.89/83.98} \\
    \cmidrule(r){2-5}
                  & LSTM (non-spiking) \cite{arjovsky2016unitary}& recurrent         & 66.5/ -       & 88.00/ - \\
                  & SRNN+ReLU (non-spiking) \cite{yin2020effective}         & recurrent         & 129.6/ -         & 93.47/ - \\
                  & GLIF* \cite{yao2022glif}             & recurrent         & 114.6/157.5         & 90.34/90.47 \\
                  & SRNN+ALIF \cite{yin2020effective,yin2021accurate}         & recurrent         & 129.6/156.3         & 91.00/94.30 \\
                  & LIF*      & recurrent         & 112.2/155.1         & 71.77/80.39 \\                                
                  & \textbf{LSTM-LIF (ours)} &  \textbf{recurrent}  &  \textbf{63.6/155.1} &  \textbf{92.69/95.07} \\
    \bottomrule
    \multirow{6}{*}{\rotatebox{90}{GSC}}   & Rate-based SNN \cite{yilmaz2020deep}    & feedforward     & 117       & 75.20 \\
                  & \textbf{LSTM-LIF (ours)} &  \textbf{feedforward} & \textbf{106.2}   & \textbf{90.57} \\
    \cmidrule(r){2-5} 
                  & SRNN+ALIF \cite{yin2021accurate}         & recurrent         & 221.7         & 92.10 \\
                  & SNN \cite{salaj2021spike}              & recurrent         & 4304.9        & 89.04 \\
                  & SNN with SFA \cite{salaj2021spike}      & recurrent         & 4307          & 91.21 \\
                  & \textbf{LSTM-LIF (ours)} &  \textbf{recurrent}  &  \textbf{196.5} &  \textbf{94.14} \\
    \bottomrule
    \multirow{9}{*}{\rotatebox{90}{SHD}}   & Feed-forward SNN \cite{cramer2020heidelberg}    & feedforward     & 108.8       & 48.60 \\
                  & \textbf{LSTM-LIF (ours)} &  \textbf{feedforward} & \textbf{108.8}   & \textbf{83.08} \\
    \cmidrule(r){2-5}
                  & SRNN \cite{cramer2020heidelberg}               & recurrent         & 108.8        & 71.4 \\
                  & Heterogeneous SRNN \cite{perez2021neural} & recurrent       & 108.8      & 82.70 \\
                  & Attention \cite{yao2021temporal}          & recurrent         & 133.8        & 81.45 \\
                  & SRNN + ALIF \cite{yin2020effective}        & recurrent         & 142.4        & 84.40 \\
                  & SRNN  \cite{zenke2021remarkable}               & recurrent         & 249        & 82.00 \\
                  & SRNN + data augm. \cite{cramer2020heidelberg}  & recurrent      & 1787.9       & 83.20 \\
                  & \textbf{LSTM-LIF (ours)} &  \textbf{recurrent}  &  \textbf{141.8} &  \textbf{88.91} \\
    \bottomrule
    \multirow{5}{*}{\rotatebox{90}{SSC}}   & Feed-forward SNN \cite{cramer2020heidelberg}    & feedforward     & 110.8       & 38.50 \\
                  & \textbf{LSTM-LIF (ours)} &  \textbf{feedforward} & \textbf{110.8}   & \textbf{63.46} \\
    \cmidrule(r){2-5}
                  & SRNN \cite{cramer2020heidelberg}                 & recurrent       & 110.8      & 50.90 \\
                  & Heterogeneous SRNN \cite{perez2021neural} & recurrent       & 110.8      & 57.3 \\
                  & \textbf{LSTM-LIF (ours)} &  \textbf{recurrent}  &  \textbf{110.8} &  \textbf{61.09} \\
    \bottomrule
    \multicolumn{5}{l}{* Our reproduced results using publicly available codes.}
  \end{tabular}}
\end{table*}

For the S-MNIST dataset, each data sample has a sequence length of 784, which requires the model to learn long-range dependencies. The LIF model performs worst on this dataset, which can be explained by the vanishing gradient problem discussed in Section \ref{sec: SNN}. As expected, the memory-augmented LSNN \cite{bellec2018long} and adaptive LIF (ALIF) \cite{yin2020effective, yin2021accurate} models achieve comparable or even better accuracies to non-spiking models, such as LSTM \cite{arjovsky2016unitary}. Our proposed LSTM-LIF model consistently outperforms these memory-augment single-compartment neuron models, suggesting its high efficacy in retaining long-term memory and handling long-term dependencies. Notably, we achieve 
99.01\% accuracy with a recurrent architecture, which is the best-reported SNN model for this dataset. The same conclusions can be drawn for the more challenging PS-MNIST dataset.

\begin{figure}[htb]
\centering
\subfloat[]{\includegraphics[width=68mm]{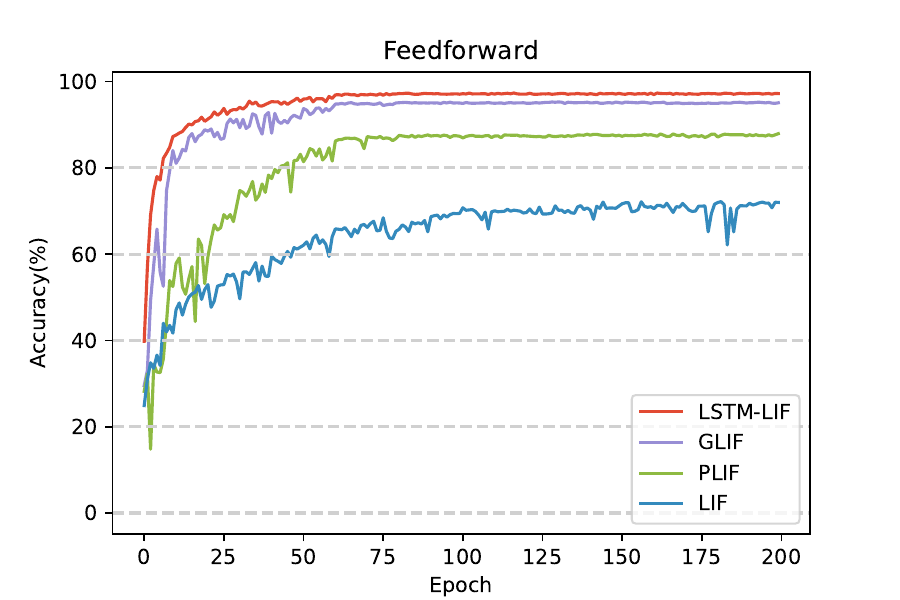}} \quad
\subfloat[]{\includegraphics[width=68mm]{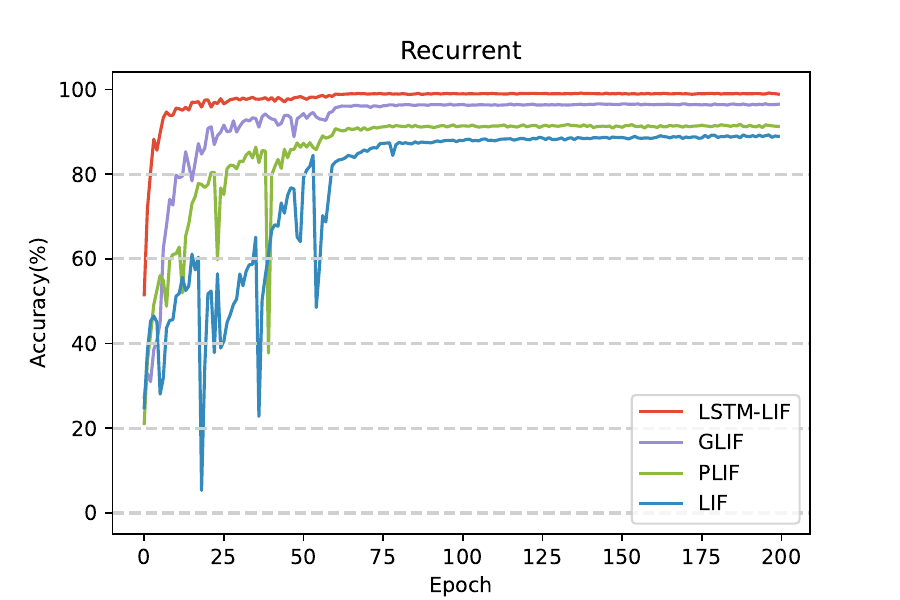}} 
\caption{Comparison of the learning curves of LIF and LSTM-LIF with (a) feedforward and (b) recurrent network architectures. }
\label{fig: lr_curve}
\end{figure}

In addition to image datasets, we further conduct experiments on speech datasets that exhibit rich temporal dynamics. For the non-spiking GSC dataset, our LSTM-LIF model achieves 90.60\% and 94.30\% accuracy for feedforward and recurrent networks respectively, surpassing SOTA models by a large margin. The SHD and SSC datasets are neuromorphic datasets that are specifically designed for benchmarking SNNs. On these datasets, our proposed LSTM-LIF exhibit a significant improvement over all other reported works.

\subsection{Rapid Learning Convergence}
\label{sec: convergence}

The gradient vanishing problem, as described in Section \ref{sec: SNN}, is notorious for BPTT training. It can result in slow convergence and unstable learning. By effectively addressing this issue, the proposed LSTM-LIF model ensures a more stable flow of gradients during the backpropagation process, leading to faster and more stable learning. 

To shed light on this, we compare the learning curve of LSTM-LIF with the LIF, GLIF, and PLIF models under the same training settings. As illustrated in Figure \ref{fig: lr_curve}, the LSTM-LIF model converges rapidly within about 25 epochs for both feedforward and recurrent networks, while the LIF model takes around 100 and 75 epochs to converge for feedforward and recurrent networks, respectively. 
Moreover, for recurrent networks, the LSTM-LIF model exhibits greater stability than the LIF and PLF models, especially during the early training stage. Although the GLIF model exhibits a similar convergence speed to the LSTM-LIF model, we notice that the LSTM-LIF model is capable of achieving higher accuracy due to the smooth loss landscape that will be explained soon.

\subsection{Stronger Network Generalization with Smooth Loss Landscape}
\label{sec: landscape}

\begin{figure}[htb]
\centering
\subfloat[]{\includegraphics[width=35mm]{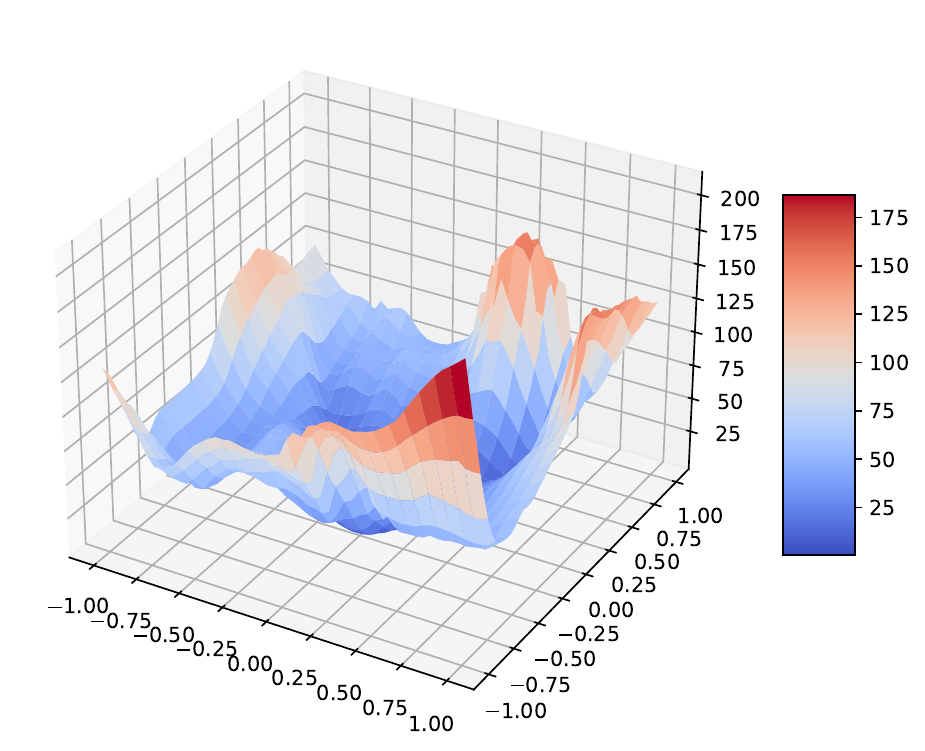}} 
\subfloat[]{\includegraphics[width=35mm]{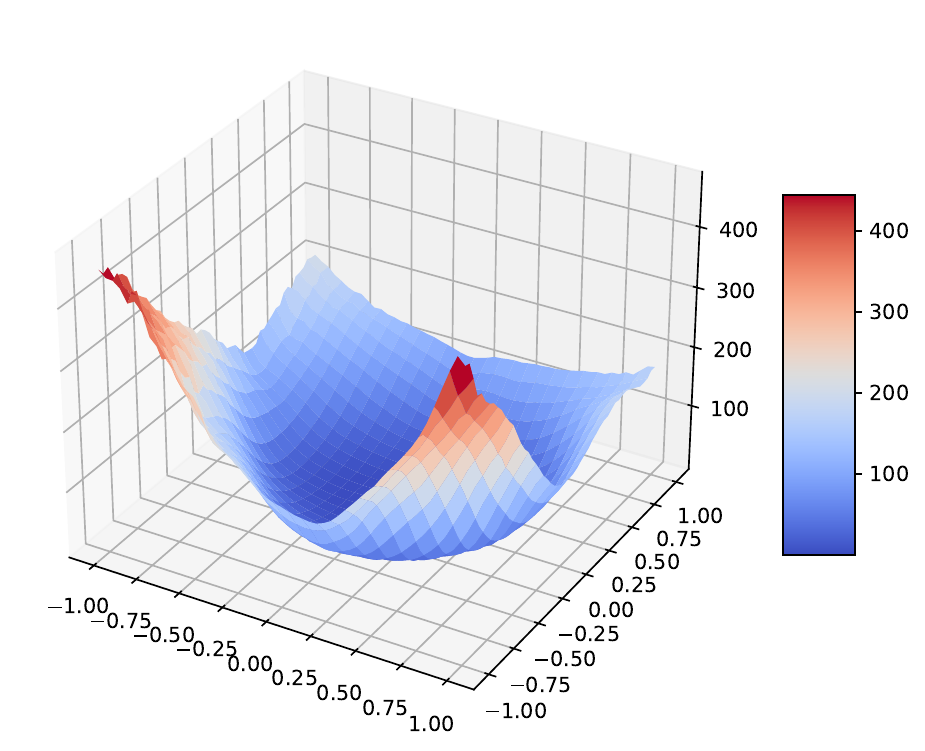}} 
\subfloat[]{\includegraphics[width=35mm]{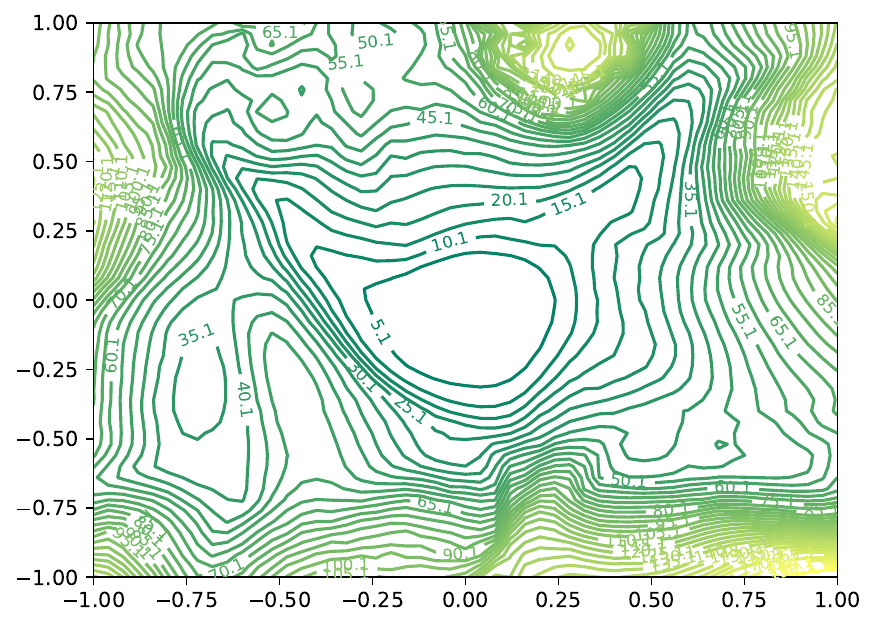}} 
\subfloat[]{\includegraphics[width=35mm]{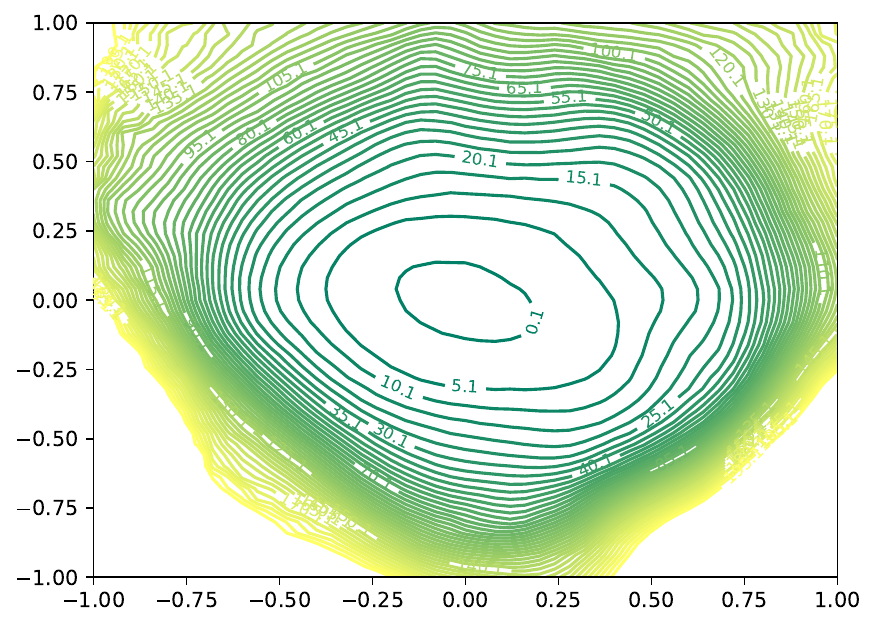}} 
\caption{Comparison of the loss landscape of (a, c) LIF and (b, d) LSTM-LIF neuron models in terms of 3D surface and 2D contour plots.}
\label{fig: loss_landscape}
\end{figure}

To investigate the reason why the LSTM-LIF model can achieve more stable learning and faster convergence than the LIF model, we further compare their loss landscape near the founded local minima.
As shown in Figure \ref{fig: loss_landscape}, it is obvious that the LSTM-LIF model exhibits a notably smoother loss landscape near the local minima compared to the LIF model. This suggests the LSTM-LIF model offers improved learning dynamics and convergence properties.
In particular, the smoother loss landscape enables a reduced likelihood of being trapped into local minima, which can lead to more stable optimization and faster convergence. Furthermore, the smoother loss landscape suggests stronger network generalization, as it is less prone to overfitting and underfitting problems. 
Overall, the observed smooth loss landscape highlights the potential of the LSTM-LIF model for more accurate and efficient learning, particularly for long temporal sequences.

\subsection{High Energy Efficiency}
\label{sec: efficient}

So far, it remains unclear whether the proposed LSTM-LIF model can make a good trade-off between model complexity and computational efficacy. To answer this question, we conduct theoretical and empirical analysis on the energy efficiency of LIF, LSTM-LIF, and non-spiking LSTM \cite{graves2012long} models. In particular, we count the accumulated (AC) and multiply-and-accumulate (MAC) operations consumed during input data processing and network update. In ANNs, the computations are all performed with MAC operations, whereas the AC operations are used predominantly in SNNs for synaptic updates. It is worth noting that the membrane potential update of spiking neurons requires several MAC operations. More detailed calculations can be found in Supplementary Materials Section \ref{app: energy_study}.

As the theoretical results presented in Table \ref{tab: energy_comp}, the energy costs of both spiking neurons (i.e., LIF and LSTM-LIF) are significantly lower than that of the LSTM model, attributed to their lesser computational complexity. Compared to the LIF model, the proposed LSTM-LIF model incurs additional $nFr_{out}E_{AC}+nE_{MAC}$ operations due to the extra computation at the dendritic compartment. 
To calculate the empirical energy cost, we perform inference on one randomly selected batch of test samples and compute the average layer-wise firing rates of these SNNs on the S-MNIST dataset. The layer-wise firing rates for LIF and LSTM-LIF models are comparable that take the values of [0.219, 0.145, 0.004] and [0.294, 0.146, 0.030], respectively. To obtain the total energy cost, we base our calculation on the 45nm CMOS process that has an estimated cost of $E_{AC}=0.9~pJ$ and $E_{MAC}=4.6~pJ$ for AC and MAC operations, respectively \cite{horowitz20141}. Despite the more complex internal structure of the proposed LSTM-LIF model, it has a comparable energy cost to the LIF model. Remarkably, our LSTM-LIF model achieves more than 100 times energy savings compared with the LSTM model, while demonstrating better temporal classification performance.

\begin{table*}[htb]
  \caption{Comparison of the theoretical and empirical energy cost of LIF, LSTM, and LSTM-LIF models. The $m$ and $n$ are the numbers of input and output neurons. $Fr_{in}$ and $Fr_{out}$ are the firing rate of input and output neurons. $E_{AC}$ and $E_{MAC}$ are the energy cost of MAC and AC operations.}
  \label{tab: energy_comp}
  \centering
  \resizebox{1.0\textwidth}{!}{%
  \begin{tabular}{lcc}
    \toprule
    \textbf{Neuron Model}   &\textbf{Theoretical Energy Cost}  &\textbf{Empirical Energy Cost (nJ)} \\
    \midrule
    LSTM            & $4(mn+nn)E_{MAC}+17nE_{MAC}$                       & 2,834.7\\
    LIF     & $mnFr_{in}E_{AC}+(nn+n)Fr_{out}E_{AC}+nE_{MAC}$    & 23.8\\
    LSTM-LIF            & $mnFr_{in}E_{AC}+(nn+2n)Fr_{out}E_{AC}+2nE_{MAC}$  & 28.2\\
                  
    \bottomrule
  \end{tabular}}
\end{table*}

\section{Conclusion}
\label{sec: conclusion}
In this paper, drawing inspiration from the multi-compartment structure of biological neurons, we proposed a novel two-compartment spiking neuron model to enhance the memory capacity of single-compartment neurons. The dendritic and somatic compartments of the proposed LSTM-LIF model are tailored to retain long-term and short-term memories, respectively. This leads to an improved ability in learning long-term dependencies. Theoretical analysis and experimental results on various temporal classification tasks demonstrate the superiority of the proposed LSTM-LIF model, including exceptional classification capability, rapid training convergence, greater network generalizability, and high energy efficiency. This work, therefore, contributes to the development of more effective and efficient spiking neurons for emerging neuromorphic computing machines. In this work, we focus our study on two-compartment neuron models, while how to generalize the design to multi-compartment neurons, with an even larger number of compartments, remains an interesting question that we will explore in future works. 


\newpage
\bibliography{citation}{}
\bibliographystyle{plain}


\newpage
\begin{center}
    \Large{\textbf{Supplementary Materials}}
\end{center}

\appendix
\section{Two-Compatment Biological Prinsky-Rinzel Neuron Model}
\label{app: method}

In this paper, we utilized the simplified Prinsky-Rinzel (P-R) neuron model proposed by Kepecs and Wang. This model represents a pyramidal cell in the CA3 region with two compartments - the somatic and dendritic compartments. The dendritic compartment is responsible for producing bursting responses, while the soma generates spikes. The somatic compartment is governed by the $I_{Na}$ and $I_{K}$ currents, whereas the dendritic compartment is characterized by the slow potassium $I_{KS}$ and a persistent sodium $I_{NaP}$ currents. The P-R neuron model consists of several parameters and two-compartment coupled equations, mathematically described by:

\begin{equation}
    \label{eq-supp: continuous_soma}
    C_{m}\frac{dV_{s}}{dt}=-I_{Na}-I_{K}-I_{Leak}+\frac{I_{link}}{P}+I_{s}
\end{equation}

\begin{equation}
    \label{eq-supp: continuous_dendrite}
    C_{m}\frac{dV_{d}}{dt}=-I_{NaP}-I_{KS}-I_{Leak}-\frac{I_{link}}{1-P}+I_{d}
\end{equation}

where $V_s$ and $V_d$ are the somatic and dendritic membrane potentials, $I_d$ and $I_s$ denote the currents applied to the soma and dendrite, respectively. Specifically, $I_s$ is assumed to be 0 in this paper and dendrite is the only part in neuron model to accept the outer currents. The membrane capacitance and the proportion of the cell area taken by soma are respectively denoted by $C_m$ and $P$.
In the Table \ref{tab: P-R_ionic_currents}, the ionic currents participate in Equation \ref{eq-supp: continuous_soma}, \ref{eq-supp: continuous_dendrite} and their corresponding calculations are presented. 
In the computations of ionic currents, $E_{Na}$, $E_{Ka}$ and $E_{L}$ represent equilibrium potentials, and $g_{Na}$, $g_{K}$, $g_{L}$, $g_{c}$, $g_{NaP}$ and $g_{KS}$ are conductances.

\begin{table}[htb]
  \caption{Calculations of ionic currents in P-R neuron model.}
  \label{tab: P-R_ionic_currents}
  \centering
  \resizebox{0.9\textwidth}{!}{%
  \begin{tabular}{lccc}
    \toprule
    \textbf{Ionic Current} &\textbf{Calculation} &\textbf{Ionic Current} &\textbf{Calculation}\\
    \midrule
    $I_{Na}$    & $g_{Na}m^3h\cdot (V_s-E_{Na})$ & $I_{link}$ & $g_{c}\cdot (V_{d}-V_{s})$     \\
    $I_{K}$     & $g_{K}n^4\cdot (V_s-E_{K})$    & $I_{NaP}$  & $g_{NaP}l^3h\cdot (V_d-E_{Na})$ \\
    $I_{Leak}$  & $g_{L}\cdot (V-E_{L})$         & $I_{KS}$   & $g_{KS}q\cdot (V_d-E_{K})$\\
                  
    \bottomrule
  \end{tabular}}
\end{table}

Following the Equations \ref{eq-supp: continuous_soma} and \ref{eq-supp: continuous_dendrite} in the continuous time, the iterative and discrete-time forms are obtained through Euler method:

\begin{equation}
    \label{eq-supp: discrete_soma}
    V_{s}[t+1]=V_{s}[t]+\frac{dt}{C_m}(-I_{Na}[t]-I_{K}[t]-I_{Leak}[t]+\frac{I_{link}[t]}{P})
\end{equation}

\begin{equation}
    \label{eq-supp: dicrete_dendrite}
    V_{d}[t+1]=V_{d}[t]+\frac{dt}{C_m}(-I_{NaP}[t]-I_{KS}[t]-I_{Leak}[t]+\frac{I_{link}[t]}{1-P}+I_{d}[t])
\end{equation}

The term $I_{link}$ encompasses the interaction between the somatic and dendritic compartments in the membrane potential. Furthermore, by incorporating the reset operation in the somatic output to transform the neuron into a spiking form, we deduce the overall dynamics of the two-compartment P-R spiking neuron model, as discussed in Section \ref{sec: PR_model}. 

\section{Experimental Details}
\label{app: exp_setting}

\subsection{Datasets}
\label{app: datasets}
In this subsection, we introduce the dataset used for this work. These datasets cover a wide range of tasks, allowing us to assess the model's capabilities in handling different types of input data.

\textbf{S-MNIST:} The Sequential-MNIST (S-MNIST) dataset is derived from the original MNIST dataset, which consists of 60,000 and 10,000 grayscale images of handwritten digits for training and testing sets with a resolution of 28 $\times$ 28 pixels. In the S-MNIST dataset, each image is converted into a vector of 784 time steps, with each pixel representing one input value at a certain time step. This dataset enables us to evaluate the performance of our model in solving sequential image classification tasks.

\textbf{PS-MNIST:} The Permuted Sequential MNIST dataset (PS-MNIST) is a variation of the Sequential MNIST dataset, in which the pixels in each image are shuffled according to a fixed random permutation. This dataset provides a more challenging task than S-MNIST, as the input sequences no longer follow the original spatial order of the images. Therefore, when learning this dataset, the model needs to capture complex, non-local, and long-term dependencies between pixels. 

\textbf{GSC:} The Google Speech Commands (GSC) has two versions, and we employ the 2nd version in this work. The GSC version 2 is a collection of 105,829 on-second-long audio clips of 35 different spoken commands, such as “yes”, “no”, “up”, “down”, “left”, “right”, etc. These audio clips are recorded by different speakers in various environments, offering a diversity of datasets to evaluate the performance of our model.

\textbf{SHD:} The Spiking Heidelberg Digits dataset is a spike-based sequence classification benchmark, consisting of spoken digits from 0 to 9 in both English and German (20 classes). The dataset contains recordings from twelve different speakers, with two of them only appearing in the test set. Each original waveform has been converted into spike trains over 700 input channels. The train set contains 8,332 examples, and the test set consists of 2,088 examples (no validation set). The SHD dataset enables us to evaluate the performance of our proposed model in processing and classifying speech data represented in spiking format.

\textbf{SSC:} The Spiking Speech Command dataset, another spike-based sequence classification benchmark, is derived from the Google Speech Commands version 2 dataset and contains 35 classes from a large number of speakers. The original waveforms have been converted to spike trains over 700 input channels. The dataset is divided into train, validation, and test splits, with 75,466, 9,981, and 20,382 examples, respectively. The SSC dataset allows us to assess the performance of our proposed spiking neuron model in processing and recognizing speech commands represented in spiking data.

\subsection{Network architecture}
\label{app: configure}
We perform experiments employing both feedforward and recurrent connection configurations. To maintain a fair comparison with existing works, we utilize network architectures exhibiting comparable parameters. These architectures and their corresponding parameters are summarized in Table \ref{tab: arch}.

\begin{table}[htb]
  \caption{Summary of network architectures and parameters.}
  \label{tab: arch}
  \centering
  \resizebox{0.8\textwidth}{!}{%
  \begin{tabular}{lccc}
    \toprule
    \textbf{Dataset}  &\textbf{Network} & \textbf{Architecture}  & \textbf{Parameters(K)}   \\
    \midrule
    \multirow{2}{*}{S-MNIST}        & feedforward   & 40-256-128-10/ 64-256-256-10      & 44.8/ 85.1    \\
                                    & recurrent     & 40-200-64-10/ 64-256-256-10      & 63.6/ 155.1    \\
    \midrule
    \multirow{2}{*}{PS-MNIST}       & feedforward   & 40-256-128-10/ 64-256-256-10      & 44.8/ 85.1    \\
                                    & recurrent     & 40-200-64-10/ 64-256-256-10      & 63.6/ 155.1    \\
    \midrule
    \multirow{2}{*}{GSC}            & feedforward   & 40-300-30-12       & 106.2    \\
                                    & recurrent     & 40-300-30-12       & 106.2    \\
    \midrule
    \multirow{2}{*}{SHD}            & feedforward   & 700-128-128-20     & 108.8    \\
                                    & recurrent     & 700-128-128-20     & 108.8    \\
    \midrule
    \multirow{2}{*}{SSC}            & feedforward   & 700-128-128-135    & 110.8    \\
                                    & recurrent     & 700-128-128-135    & 110.8    \\
    \bottomrule
  \end{tabular}}
\end{table}

\subsection{LSTM-LIF model hyper-parameters}
\label{app: hp}
In this section, we provide our detailed settings on the hyper-parameters of LSTM-LIF neuron model in Table \ref{tab: hp}, including the $\gamma$, initial values of $\beta$ and neuronal threshold $\mathcal{V}_{th}$.

\begin{table}[htb]
  \caption{Training hyper-parameters for LSTM-LIF.}
  \label{tab: hp}
  \centering
  \resizebox{0.5\textwidth}{!}{%
  \begin{tabular}{lcccc}
    \toprule
    \textbf{Dataset}  &\textbf{Network} & \textbf{$\gamma$}  & \textbf{$\beta$}  & $\mathcal{V}_{th}$ \\
    \midrule
    \multirow{2}{*}{S-MNIST}        & feedforward   & 0.5      & (-0.5, 0.5)   & 1.0  \\
                                    & recurrent     & 0.5      & (-0.5, 0.5)   & 1.0 \\
    \midrule
    \multirow{2}{*}{PS-MNIST}       & feedforward   & 0.7      & (-0.5, 0.5)   & 1.5 \\
                                    & recurrent     & 1.0      & (-0.5, 0.5)   & 1.8 \\
    \midrule
    \multirow{2}{*}{GSC}            & feedforward   & 0.6       & (-0.5, 0.5)  & 1.2  \\
                                    & recurrent     & 0.6       & (-0.5, 0.5)  & 1.3  \\
    \midrule
    \multirow{2}{*}{SHD}            & feedforward   & 0.5    & (-0.5, 0.5)  & 1.5  \\
                                    & recurrent     & 0.5   & (-0.5, 0.5)  & 1.5  \\
    \midrule
    \multirow{2}{*}{SSC}            & feedforward   & 0.5    & (-0.5, 0.5)  & 1.5  \\
                                    & recurrent     & 0.5    & (-0.5, 0.5)  & 1.5  \\
    \bottomrule
  \end{tabular}}
\end{table}

\subsection{Training configuration}
\label{app: config}
We train the S-MNIST and PS-MNIST datasets for 200 epochs utilizing the Adam optimizer. Their initial learning rates are set to 0.0005 for both feedforward and recurrent networks with the learning rates decaying by a factor of 10 at epochs 60 and 80. For the GSC, SHD, and SSC datasets, we train the models for 100 epochs using the Adam optimizer. 
The initial learning rate of GSC datasets is 0.001 for both feedforward and recurrent networks with the decaying by 10 at epochs 60, 90, and 120.
The initial learning rate is set to 0.0005, and 0.005 for feedforward and recurrent networks on the SHD dataset, with the learning rate decaying to 0.8 times its previous value every 10 epochs. For the SSC dataset, the initial learning rates are 0.0001 for both feedforward and recurrent networks, and decay to 0.8 times their previous values every 10 epochs. We train S-MNIST, PS-MNIST, and GSC tasks on Nvidia Geforce GTX 3090Ti GPUs with 24GB memory, and train SHD and SSC tasks on Nvidia Geforce GTX 1080Ti GPUs with 12GB memory. 

\subsection{Source Code}
All codes to reproduce our results will be released after the reviewing process.

\section{Study on energy efficiency}
\label{app: energy_study}
We formulate the theoretical energy cost for LSTM, LIF, and LSTM-LIF recurrent networks based on their computational dynamics calculations.
Table \ref{tab: energy cost} presents the detailed calculation of theoretical energy cost for each model. 

\begin{table}[htb]
  \caption{Computations on the energy cost of LIF, LSTM-LIF, and LSTM.}
  \label{tab: energy cost}
  \centering
  \resizebox{1.0\textwidth}{!}{%
  \begin{tabular}{lccc}
    \toprule
    \textbf{Neuron Model}  &\textbf{Dynamics} & \textbf{Step Cost}  & \textbf{Total Cost}   \\
    \midrule
    \multirow{2}{*}{LIF}            & $\mathcal{I}_{t}=\mathcal{W}^{m,n}X^m+\mathcal{W}^{n,n}\mathcal{S}^n_{t-1}$     &      $(mnFr_{in}+nnFr_{out})E_{AC}$      & $mnFr_{in}E_{AC}+(nn+n)Fr_{out}E_{AC}$    \\
                                    & $\mathcal{U}_{t}=\beta\mathcal{U}_{t-1}+\mathcal{I}_{t}-\mathcal{V}_{th}\mathcal{S}^n_{t-1}$     & $nFr_{out}E_{AC}+nE_{MAC}$      & $+nE_{MAC}$    \\
    \midrule
    \multirow{3}{*}{LSTM-LIF}       & $\mathcal{I}_{t}=\mathcal{W}^{m,n}X^m+\mathcal{W}^{n,n}\mathcal{S}^n_{t-1}$ & $(mnFr_{in}+nnFr_{out})E_{AC}$      & $mnFr_{in}E_{AC}$       \\
                                    & $\mathcal{U}^{D}_{t}=\mathcal{U}^{D}_{t-1}+\mathcal{I}_{t}+\beta_1\mathcal{U}^{S}_{t-1}-\gamma\mathcal{S}^n_{t-1}$  & $nFr_{out}E_{AC}+nE_{MAC}$      & $+(nn+2n)Fr_{out}E_{AC}$       \\
                                    & $\mathcal{U}^{S}_{t}=\mathcal{U}^{S}_{t-1}+\beta_2\mathcal{U}^{D}_{t}-\mathcal{V}_{th}\mathcal{S}^n_{t-1}$& $nFr_{out}E_{AC}+nE_{MAC}$&$+2nE_{MAC}$ \\
    \midrule
    \multirow{6}{*}{LSTM}           & $f_{t}=\sigma_g(\mathcal{W}_{f}x_{t}+{U}_{f}h_{t-1}+b_{f})$        & $n(m+n+2)E_{MAC}$        &                      \\
                                    & $i_{t}=\sigma_g(\mathcal{W}_{i}x_{t}+{U}_{i}h_{t-1}+b_{i})$        & $n(m+n+2)E_{MAC}$        &                      \\
                                    & $o_{t}=\sigma_g(\mathcal{W}_{o}x_{t}+{U}_{o}h_{t-1}+b_{o})$        & $n(m+n+2)E_{MAC}$        & $4(mn+nn)E_{MAC}$    \\
                                    & $\hat{c}_{t}=\sigma_c(\mathcal{W}_{c}x_{t}+{U}_{c}h_{t-1}+b_{c})$  & $n(m+n+4)E_{MAC}$        & $17nE_{MAC}$         \\
                                    & $c_{t}=f_{t} \odot c_{t-1}+i_{t} \odot \hat{c}_{t}$                & $2nE_{MAC}$              &                      \\
                                    & $h_{t}=o_{t} \odot \sigma_{h}(c_{t})$                              & $5nE_{MAC}$              &                      \\

    \bottomrule
  \end{tabular}}
\end{table}

\end{document}